\documentclass[11pt]{article}

\usepackage[final]{acl}

\usepackage{times}
\usepackage{latexsym}
\usepackage[T1]{fontenc}
\usepackage[utf8]{inputenc}
\usepackage{microtype}
\usepackage{graphicx}
\usepackage{amsmath}
\usepackage{amssymb}
\usepackage{amsthm}

\usepackage{booktabs}
\usepackage{algorithm}
\usepackage{algpseudocode}
\usepackage{tikz}
\usepackage{pgfplots}
\usepackage{adjustbox}
\pgfplotsset{compat=1.17}
\usetikzlibrary{shapes,arrows,positioning,fit,calc}
\usepackage[section]{placeins}

\newcommand{\speedup}{1.80}
\newcommand{\speedupx}{1.80$\times$}
\newcommand{\stsb}{0.763}

\newcommand{\avgexitlayer}{6.7}

\newcommand{\exitlsix}{38.9}
\newcommand{\exitlseven}{91.9}
\newcommand{\exitleight}{97.6}
\newcommand{\exitlnine}{98.1}
\newcommand{\exitlten}{99.5}
\newcommand{\exitleleven}{100.0}

\newcommand{\simlsix}{0.945}
\newcommand{\simlseven}{0.963}
\newcommand{\simleight}{0.968}
\newcommand{\simlnine}{0.970}
\newcommand{\simlten}{0.975}
\newcommand{\simleleven}{0.993}

\newcommand{\nnlsix}{0.79}
\newcommand{\nnlseven}{0.84}
\newcommand{\nnleight}{0.86}
\newcommand{\nnlnine}{0.86}
\newcommand{\nnlten}{0.87}
\newcommand{\nnleleven}{0.93}

\providecommand{\monitoringoverhead}{22.2}

\providecommand{\mediancontraction}{0.911}

\providecommand{\leapsimi}{0.769}

\providecommand{\leapstabii}{0.931}

\providecommand{\leapsimv}{0.919}

\providecommand{\leapsimvii}{0.963}

\providecommand{\leapsimxii}{1.000}
\providecommand{\leapstabxii}{0.994}

\providecommand{\firstviablelayer}{7}
\providecommand{\avglatestability}{0.981}

\providecommand{\layertwelvealignment}{1.000}
\providecommand{\avglatealignment}{0.640}

\providecommand{\nnfailpctliv}{41.8}

\providecommand{\nnfailpctlvii}{0.6}

\providecommand{\nnfailpctlviii}{0.2}

\providecommand{\stsbSpearmanWithCI}{0.760 $\pm$ 0.006}

\providecommand{\stsbbaseline}{0.777}

\providecommand{\eesimlsix}{0.945}
\providecommand{\eeexitsix}{38.9}
\providecommand{\basesimlsix}{0.162}
\providecommand{\baseexitsix}{0.0}

\providecommand{\eesimlseven}{0.963}
\providecommand{\eeexitseven}{91.9}
\providecommand{\basesimlseven}{0.215}
\providecommand{\baseexitseven}{0.0}

\providecommand{\eesimleight}{0.968}
\providecommand{\eeexiteight}{97.6}
\providecommand{\basesimleight}{0.285}
\providecommand{\baseexiteight}{0.0}

\providecommand{\eesimlnine}{0.970}
\providecommand{\eeexitnine}{98.1}
\providecommand{\basesimlnine}{0.374}
\providecommand{\baseexitnine}{0.0}

\providecommand{\eesimlten}{0.975}
\providecommand{\eeexitten}{99.5}
\providecommand{\basesimlten}{0.547}
\providecommand{\baseexitten}{0.0}

\providecommand{\eesimleleven}{0.994}
\providecommand{\eeexiteleven}{100.0}
\providecommand{\basesimleleven}{0.858}
\providecommand{\baseexiteleven}{0.0}

\providecommand{\eesimltwelve}{1.000}
\providecommand{\eeexittwelve}{100.0}
\providecommand{\basesimltwelve}{1.000}
\providecommand{\baseexittwelve}{100.0}

\newcommand{\latencyFullMs}{8.46}
\newcommand{\latencyEEMs}{5.25}
\newcommand{\latencySpeedup}{1.61}
\newcommand{\latencyEfficiency}{80.6}

\newcommand{\latencyFullMsEight}{11.51}
\newcommand{\latencyEEMsEight}{8.75}
\newcommand{\latencySpeedupEight}{1.32}

\newcommand{\latencyFullMsThirtytwo}{13.14}
\newcommand{\latencyEEMsThirtytwo}{10.61}
\newcommand{\latencySpeedupThirtytwo}{1.24}

\title{LEAP: Layer-wise Exit-Aware Pretraining for Efficient Transformer Inference}

\author{
  Shashank Kapadia$^{1}$, Deep Naryan Mishra$^{1}$, Sujal Reddy Alugubelli$^{2}$, \\
  Haoan Wang$^{1}$, Saipraveen Vabbilisetty$^{1}$, Rishi Bhatia$^{1}$, Anupriya Sharma$^{1}$ \\[2pt]
  $^{1}$Walmart Inc.\quad $^{2}$Sam's Club \\[2pt]
  \texttt{\{Shashank.Kapadia, Deep.Mishra, Haoan.Wang\}@walmart.com} \\
  \texttt{\{Saipraveen.Vabbilisetty, Rishi.Bhatia, Anupriya.Sharma\}@walmart.com} \\
  \texttt{SujalReddy.Alugubelli@samsclub.com}
}

\begin{document}
\maketitle

\begin{abstract}
Layer-aligned distillation and convergence-based early exit represent two predominant computational efficiency paradigms for transformer inference; yet we establish that they exhibit systematic incompatibility under standard deployment conditions for convergence-based early exit. Distillation objectives that align intermediate student layers to teacher representations suppress the representational convergence that early-exit mechanisms exploit, rendering such mechanisms ineffective on distilled models.

We introduce \textbf{LEAP} (Layer-wise Exit-Aware Pretraining), an auxiliary training objective that reconciles this incompatibility. LEAP requires no architectural modifications; it augments standard distillation with a single constraint ensuring intermediate layers approximate final-layer representations. LEAP-MiniLM achieves \textbf{\latencySpeedup$\times$ measured wall-clock speedup} (batch=1, NVIDIA L4) at $\theta$=0.95, with 91.9\% of samples exiting by layer 7 and \speedupx{} theoretical layer reduction, where standard distilled models achieve \emph{zero} effective speedup. We validate across sentence similarity (STS-B: \stsbSpearmanWithCI) and retrieval benchmarks (BEIR), providing operational guidance including latency measurements, decision thresholds, and deployment criteria.

\noindent\textbf{Industry Track: Emerging}
\end{abstract}

%==============================================================================
% SECTION 1: INTRODUCTION
%==============================================================================
\section{Introduction}

Dense text embeddings underpin modern information retrieval~\citep{reimers-gurevych-2019-sentence}, semantic search, and recommendation systems. Two predominant paradigms exist for efficient embedding inference: \emph{knowledge distillation}, producing compact models such as MiniLM~\citep{wang2020minilm} and DistilBERT~\citep{sanh2019distilbert}, and \emph{early exit}, terminating computation when representations converge~\citep{xin-etal-2020-deebert,zhou-etal-2020-bert}. These orthogonal optimizations should, in principle, compose - distilled models should admit further acceleration through early exit. We establish that this assumption does not hold empirically.

\subsection{The Distillation-Exit Incompatibility}

The incompatibility manifests operationally: practitioners deploy distilled models with early-exit infrastructure, observe termination conditions satisfied at intermediate layers, yet measure \emph{zero} effective layer reduction. Per-layer convergence monitoring imposes overhead without compensating early terminations, causing net throughput degradation.

\begin{figure}[t]
\centering
\includegraphics[width=\columnwidth]{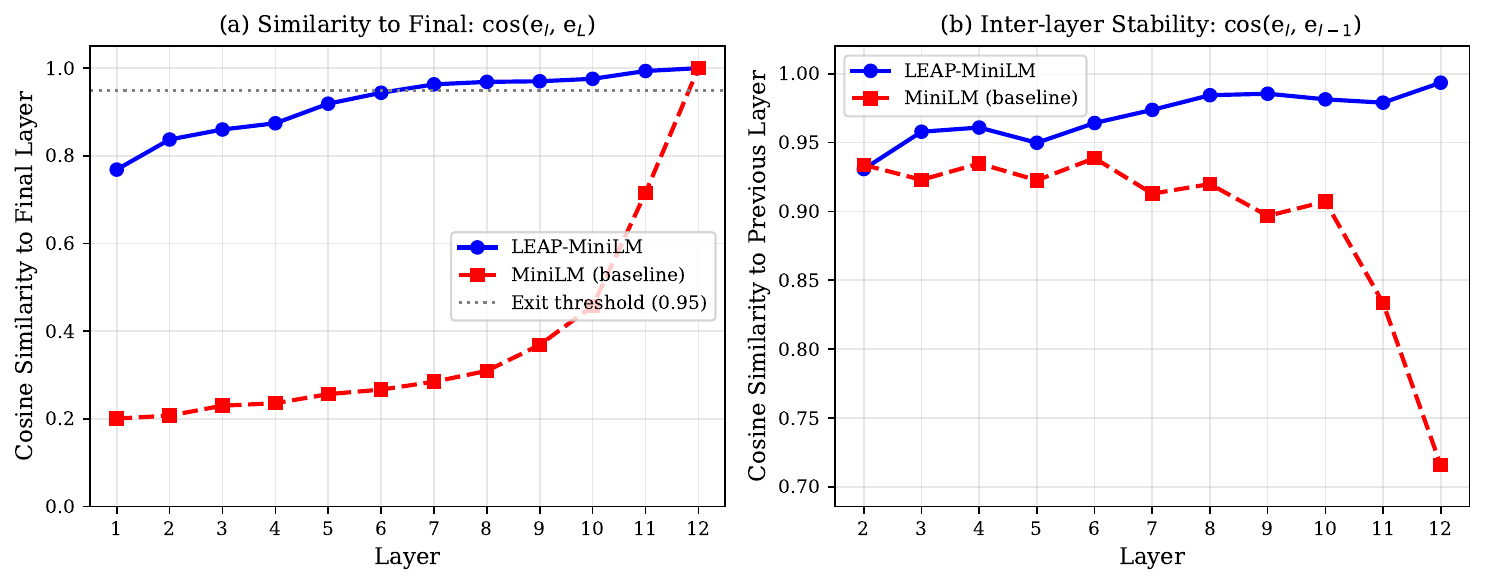}
\caption{Layer dynamics comparison. (a)~Cosine similarity to final layer: LEAP-MiniLM exceeds exit threshold ($\theta$=0.95) from layer \firstviablelayer{}; baseline MiniLM remains below threshold until layer 12. (b)~Inter-layer stability: LEAP maintains high stability (avg: \avglatestability{}) while achieving early convergence.}
\label{fig:layer-dynamics}
\end{figure}

The underlying cause is intrinsic to the optimization objective, not hyperparameter selection. Layer-aligned distillation objectives suppress the representational redundancy that early exit exploits. Standard distilled models exhibit uniformly distributed computation across layers - representations never stabilize before the final layer (Figure~\ref{fig:layer-dynamics}).

\subsection{The Mechanism: Why Layer-Aligned Distillation Eliminates Exit Points}

Layer-aligned distillation trains each student layer to match the corresponding teacher layer, inducing uniform distribution of the teacher's representational capacity across the student's depth hierarchy. This suppresses monotonic convergence---later layers performing diminishing transformations---precisely the property early exit requires.

\subsection{LEAP: Reconciling Distillation with Early Exit}

We introduce \textbf{LEAP} (Layer-wise Exit-Aware Pretraining), an auxiliary training objective that reconciles this incompatibility. LEAP augments standard distillation with a single constraint: ensuring intermediate layers produce representations approximating the final layer. This objective requires no architectural modifications and introduces no inference-time parameters.

\subsection{Contributions}

\begin{enumerate}
\item \textbf{Empirical characterization}: We establish that layer-aligned distillation and convergence-based early exit exhibit systematic incompatibility, characterizing conditions under which this conflict arises (\S\ref{sec:analysis}).
\item \textbf{LEAP}: An auxiliary training objective that reconciles distillation with early exit, requiring no architectural modifications.
\item \textbf{Empirical validation}: \latencySpeedup$\times$ wall-clock speedup (\speedupx{} layer reduction) where standard distillation yields none, validated on sentence similarity and retrieval tasks with GPU latency benchmarks across batch sizes.
\item \textbf{Operational framework}: Decision thresholds, wall-clock measurements, failure diagnostics, and deployment criteria for practitioners.
\end{enumerate}

%==============================================================================
% SECTION 2: RELATED WORK (0.5 pages)
%==============================================================================
\section{Related Work}

\paragraph{Early Exit.}
DeeBERT~\citep{xin-etal-2020-deebert}, FastBERT~\citep{liu-etal-2020-fastbert}, PABEE~\citep{zhou-etal-2020-bert}, LeeBERT~\citep{zhu-2021-leebert}, BERxiT~\citep{xin-etal-2021-berxit}, and CALM~\citep{schuster2022confident} enable adaptive inference via learned exit classifiers or patience-based criteria. DeeBERT and LeeBERT attach trained exit heads at each layer, adding parameters and requiring task-specific fine-tuning of these heads. PABEE and BERxiT employ patience counters or learned entropy thresholds. All implicitly presuppose sufficient inter-layer representational redundancy - an assumption that fails for distilled models. LEAP's convergence-based criterion is parameter-free and task-agnostic: it requires no additional modules, operating directly on the similarity between intermediate and final representations. We find that DeeBERT-style learned exit heads achieve only 0.26 STS-B Spearman on our MiniLM-L12 backbone (vs.\ LEAP's 0.76), as these heads are designed for classification confidence, not embedding quality.

\paragraph{Knowledge Distillation.}
MiniLM~\citep{wang2020minilm}, DistilBERT~\citep{sanh2019distilbert}, and TinyBERT~\citep{jiao-etal-2020-tinybert} compress transformers via layer-aligned objectives. DynaBERT~\citep{hou2020dynabert} enables dynamic width/depth. Critically, standard distillation distributes computation uniformly across layers, eliminating the redundancy early exit exploits.

\paragraph{Efficient Inference \& Dynamic Computation.}
Pruning~\citep{michel2019sixteen}, quantization~\citep{zafrir2019q8bert}, and LayerDrop~\citep{fan2019reducing} reduce per-layer cost; ACT~\citep{graves2016adaptive}, Universal Transformers~\citep{dehghani2018universal}, and PonderNet~\citep{banino2021pondernet} learn halting criteria. Matryoshka representations~\citep{kusupati2022matryoshka} enable variable embedding dimensions, reducing memory and distance computation costs. LEAP is orthogonal: Matryoshka varies the \emph{width} of the final embedding while LEAP varies the \emph{depth} at which it is computed. Combined, this could yield multiplicative benefits---Matryoshka for storage/retrieval efficiency, LEAP for inference latency.

%==============================================================================
% SECTION 3: METHODOLOGY (1.0 page)
%==============================================================================
\section{Methodology}

\subsection{Problem Analysis: Why Layer-Aligned Distillation Suppresses Early Exit}
\label{sec:analysis}

Consider a teacher model $M_t$ with $L_t$ layers and a student $M_s$ with $L_s < L_t$ layers. Standard distillation minimizes:
\begin{equation}
\mathcal{L}_{\text{distill}} = \sum_{l=1}^{L_s} \text{KL}(\mathbf{h}_s^{(l)} \| \mathbf{h}_t^{(\pi(l))})
\end{equation}
where $\pi(l)$ maps student layers to teacher layers and KL denotes Kullback-Leibler divergence over attention distributions (as in MiniLM). LEAP instead employs cosine similarity for embedding alignment (\S\ref{sec:leap}). This formulation explicitly trains \emph{every} student layer to match the teacher, distributing computation uniformly.

\paragraph{The Redundancy Problem.}
Early-exit predicates necessitate monotonically diminishing inter-layer transformations - i.e., $\|\mathbf{p}_{l+1} - \mathbf{p}_l\| \to 0$ as $l \to L$. Define the \emph{contraction ratio}:
\begin{equation}
\gamma_l = \frac{\|\mathbf{p}_{l+1} - \mathbf{p}_l\|}{\|\mathbf{p}_l - \mathbf{p}_{l-1}\|}
\end{equation}
where $\mathbf{p}_l$ is the mean-pooled representation at layer $l$. Early exit is viable when $\gamma_l < 1$ consistently. Standard distillation produces $\gamma_l \approx 1$ throughout - no natural exit points.

\paragraph{Scope.} This analysis applies to layer-aligned distillation with uniform or proportional layer mapping $\pi$. Distillation variants that match only the final output (e.g., output-only KD) do not impose per-layer alignment and may preserve intermediate convergence. We focus on layer-aligned objectives as they constitute standard practice in MiniLM, TinyBERT, and similar widely deployed models.

\subsection{LEAP: Preserving Early Exit During Distillation}
\label{sec:leap}

We introduce \textbf{LEAP}, an auxiliary training objective that achieves distillation benefits while preserving early-exit capability:
\begin{equation}
\mathcal{L}_{\text{LEAP}} = \mathcal{L}_{\text{final}} + \alpha \mathcal{L}_{\text{inter}} + \beta \mathcal{L}_{\text{exit}} + \delta \mathcal{L}_{\text{contrast}}
\end{equation}
The principal methodological contribution is that $\mathcal{L}_{\text{exit}}$ is essential - without it, distilled models cannot support early exit (see Appendix~\ref{sec:appendix-ablation}).

\paragraph{Final Layer Distillation ($\mathcal{L}_{\text{final}}$).}
Standard output matching: $\mathcal{L}_{\text{final}} = 1 - \cos(\mathbf{e}_s^{(L_s)}, \mathbf{e}_t^{(L_t)})$ where $\mathbf{e}$ denotes the normalized mean-pooled embedding.

\paragraph{Intermediate Layer Distillation ($\mathcal{L}_{\text{inter}}$).}
Layer-wise matching: $\mathcal{L}_{\text{inter}} = \frac{1}{L_s} \sum_{l=1}^{L_s} (1 - \cos(\mathbf{e}_s^{(l)}, \mathbf{e}_t^{(\pi(l))}))$ with uniform layer mapping.

\paragraph{Early Exit Quality Loss ($\mathcal{L}_{\text{exit}}$).}
The key innovation: we train intermediate layers to produce valid exit points employing a \emph{soft margin loss} with dual targets - the teacher's final layer and the student's own final layer:
\begin{align}
\mathcal{L}_{\text{exit}}^{(t)} &= \frac{1}{L_s} \sum_{l=1}^{L_s} w_l \cdot \sigma\bigl(10 \cdot (\tau \notag \\
  &\quad - \cos(\mathbf{e}_s^{(l)}, \mathbf{e}_t^{(L_t)}))\bigr) \\
\mathcal{L}_{\text{exit}}^{(s)} &= \frac{1}{L_s-1} \sum_{l=1}^{L_s-1} w_l \cdot \sigma\bigl(10 \cdot (\tau \notag \\
  &\quad - \cos(\mathbf{e}_s^{(l)}, \text{sg}(\mathbf{e}_s^{(L_s)})))\bigr)
\end{align}
where $\sigma$ is sigmoid, $\tau=0.98$ is the exit threshold, $w_l$ are layer weights, and $\text{sg}(\cdot)$ denotes stop-gradient. Total: $\mathcal{L}_{\text{exit}} = \mathcal{L}_{\text{exit}}^{(t)} + 0.7 \mathcal{L}_{\text{exit}}^{(s)}$; the 0.7 weight balances teacher-aligned quality with inference-time self-consistency.

\paragraph{Contrastive Distillation ($\mathcal{L}_{\text{contrast}}$).}
We align similarity structure: $\mathcal{L}_{\text{contrast}} = \text{KL}(\text{softmax}(\mathbf{S}_s/\tau_c) \| \text{softmax}(\mathbf{S}_t/\tau_c))$ where $\mathbf{S}$ are batch-wise similarity matrices.

\subsection{Convergence-Based Early Exit Inference}
\label{sec:early-exit}

At inference, we employ $k$-skip patience-based early exit requiring no additional parameters. The patience parameter $k$ specifies the layer gap over which convergence is measured - comparing the current layer's representation to that of $k$ layers prior. After each layer $l \geq l_{\min}$, we compute mean-pooled representations $\mathbf{p}_l$ and exit when:
\begin{equation}
s_l = \cos(\mathbf{p}_l, \mathbf{p}_{l-k}) \geq \theta
\end{equation}
where $k$ is patience (default $k$=1), $l_{\min}$=6, and $\theta$=0.95. Samples meeting the convergence criterion exit immediately; remaining samples proceed to deeper layers. Full pseudocode in Appendix~\ref{sec:appendix-algorithm}.

\paragraph{Training vs.\ Inference Threshold.} We use different thresholds: $\tau{=}0.98$ (training) pushes representations closer to final layer; $\theta{=}0.95$ (inference) allows earlier exits with headroom. This separation provides robustness to distribution shift.

%==============================================================================
% SECTION 4: RESULTS (2.0 pages - expanded with more tables/figures)
%==============================================================================
\section{Results}
\label{sec:results}

\subsection{Experimental Setup}

We compare two 12-layer models: (1)~\textbf{MiniLM-L12 (baseline)}: trained with our distillation pipeline \emph{without} $\mathcal{L}_{\text{exit}}$; (2)~\textbf{LEAP-MiniLM-L12}: trained with full LEAP. Both use the same architecture, training data (AllNLI, 1.5M sentences), and teacher (\texttt{bert-large-nli-mean-tokens}). This controlled comparison isolates the exit constraint effect. Training: 10 epochs, batch size 64, LR $5 \times 10^{-5}$, $\tau=0.98$ (training), $\theta=0.95$ (inference). Full details in Appendix~\ref{sec:appendix-setup}.

\subsection{Main Results}

Table~\ref{tab:main-results} presents our main findings. Conventionally distilled models exhibit zero marginal utility from early-exit mechanisms - representations never stabilize before the final layer. LEAP-MiniLM achieves \latencySpeedup$\times$ wall-clock speedup (\speedupx{} layer reduction) with \exitlseven\% of samples exiting by layer 7.

% Main results table with wall-clock validation
\begin{table}[t]
\centering
\footnotesize
\setlength{\tabcolsep}{2pt}
\begin{adjustbox}{max width=\columnwidth}
\begin{tabular}{lccccc}
\toprule
\textbf{Model} & \textbf{STS-B} $\rho$ & \textbf{Layer Red.} & \textbf{Wall-Clock}$^\dagger$ & \textbf{E[layers]} & \textbf{Exit@L7} \\
\midrule
Published MiniLM-L12-v2$^\ddagger$ & 0.831 & 1.00$\times$ & 1.00$\times$ & 12.0 & 0\% \\
MiniLM-L12 (baseline) & \stsbbaseline & 1.00$\times$ & 1.00$\times$ & 12.0 & 0\% \\
LEAP-MiniLM-L12 & \stsbSpearmanWithCI & \speedupx & \latencySpeedup$\times$ & \avgexitlayer & \exitlseven\% \\
\bottomrule
\end{tabular}
\end{adjustbox}
\caption{Main results at $\theta$=0.95. STS-B with 95\% CI (3 seeds). Layer Red.\ = $L/\mathbb{E}[\text{exit layer}]$. $^\dagger$Wall-clock speedup measured on NVIDIA L4, batch=1. Baseline trained with our pipeline without $\mathcal{L}_{\text{exit}}$. $^\ddagger$External: sentence-transformers/all-MiniLM-L12-v2.}
\label{tab:main-results}
\end{table}

\paragraph{External Validation.} We verify generalization beyond our controlled setup using the published \texttt{all-MiniLM-L12-v2} (STS-B: 0.831). This production model achieves \textbf{0\% exit rate} at all intermediate layers (L7 similarity: 0.29 vs.\ LEAP's 0.96), confirming the incompatibility is intrinsic to layer-aligned distillation, not our training pipeline.

\paragraph{Cross-Distillation Generalization.} We test exit compatibility across distillation methods (Table~\ref{tab:cross-distill}). Both layer-aligned variants---TinyBERT (MSE on hidden states) and MiniLM-L6 (KL on attention)---exhibit near-zero convergence-based exit rates. DistilBERT, which uses output-only distillation without per-layer alignment, naturally preserves convergence (71.5\% exit rate). This confirms per-layer alignment is the root cause.

\begin{table}[t]
\centering
\small
\begin{tabular}{@{}llc@{}}
\toprule
\textbf{Model} & \textbf{Distillation} & \textbf{Max Exit Rate} \\
\midrule
TinyBERT-6 & Layer-aligned (MSE) & \textbf{0.0\%} \\
MiniLM-L6-v2 & Layer-aligned (KL) & \textbf{0.7\%} \\
DistilBERT-6 & Output-only & 71.5\% \\
\bottomrule
\end{tabular}
\caption{Exit compatibility across distillation methods. Layer-aligned methods suppress early exit; output-only distillation preserves it.}
\label{tab:cross-distill}
\end{table}

\subsection{Layer-wise Analysis: Why LEAP Enables Early Exit}

Table~\ref{tab:layer-comparison} reveals the key difference between standard distillation and LEAP. Standard MiniLM achieves low similarity to the final layer until the very end (\basesimlsix{} at L6, only reaching 1.0 at L12), resulting in 0\% exit rate at all intermediate layers. LEAP-MiniLM achieves high similarity early (\eesimlsix{} at L6, \eesimlseven{} at L7), enabling \eeexitseven\% cumulative exit by layer 7.

% Auto-generated comparison table
\begin{table}[t]
\centering
\small
\setlength{\tabcolsep}{4pt}
\begin{tabular}{@{}lcccc@{}}
\toprule
\textbf{Layer} & \multicolumn{2}{c}{\textbf{MiniLM (baseline)}} & \multicolumn{2}{c}{\textbf{LEAP-MiniLM (ours)}} \\
\cmidrule(lr){2-3} \cmidrule(lr){4-5}
 & Sim & Exit\% & Sim & Exit\% \\
\midrule
6  & \basesimlsix     & \baseexitsix\%     & \eesimlsix     & \eeexitsix\% \\
7  & \basesimlseven   & \baseexitseven\%   & \eesimlseven   & \eeexitseven\% \\
8  & \basesimleight   & \baseexiteight\%   & \eesimleight   & \eeexiteight\% \\
9  & \basesimlnine    & \baseexitnine\%    & \eesimlnine    & \eeexitnine\% \\
10 & \basesimlten     & \baseexitten\%     & \eesimlten     & \eeexitten\% \\
11 & \basesimleleven  & \baseexiteleven\%  & \eesimleleven  & \eeexiteleven\% \\
12 & \basesimltwelve  & \baseexittwelve\%  & \eesimltwelve  & \eeexittwelve\% \\
\bottomrule
\end{tabular}
\caption{Layer-wise comparison: cosine similarity to final layer and exit rate at $\theta$=0.95.}
\label{tab:layer-comparison}
\end{table}

Table~\ref{tab:layer-analysis} provides detailed per-layer analysis of LEAP-MiniLM, showing similarity to final layer, cumulative exit rate, and NN@10 retrieval overlap. The ``Viable'' column marks layers where early-exit quality remains acceptable ($\geq$0.95 similarity and $\geq$0.80 NN@10).

% Auto-generated layer analysis table
\begin{table}[t]
\centering
\small
\begin{tabular}{ccccc}
\toprule
\textbf{Layer} & \textbf{Similarity} & \textbf{Exit Rate} & \textbf{NN@10} & \textbf{Viable} \\
\midrule
6  & \simlsix     & \exitlsix\%     & \nnlsix  & $\times$ \\
7  & \simlseven   & \exitlseven\%   & \nnlseven & \checkmark \\
8  & \simleight   & \exitleight\%   & \nnleight & \checkmark \\
9  & \simlnine    & \exitlnine\%    & \nnlnine  & \checkmark \\
10 & \simlten     & \exitlten\%     & \nnlten   & \checkmark \\
11 & \simleleven  & \exitleleven\%  & \nnleleven & \checkmark \\
\bottomrule
\end{tabular}
\caption{Per-layer early exit analysis at $\theta=0.95$. Viable = similarity $\geq$0.95 \emph{and} NN@10 $\geq$0.80. Layer 7 is the first viable exit point.}
\label{tab:layer-analysis}
\end{table}

\subsection{Exit Distribution}

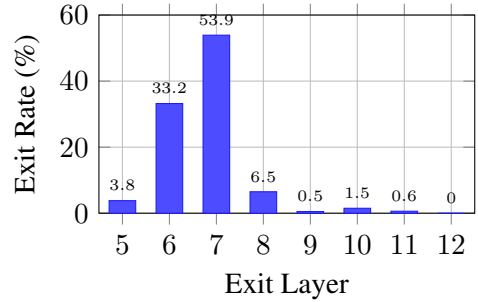
\begin{figure}[t]
\centering
\begin{tikzpicture}
\begin{axis}[
    width=0.85\columnwidth,
    height=4.2cm,
    xlabel={Exit Layer},
    ylabel={Exit Rate (\%)},
    ybar,
    bar width=10pt,
    xmin=4.5, xmax=12.5,
    ymin=0, ymax=60,
    grid=major,
    xtick={5,6,7,8,9,10,11,12},
    nodes near coords,
    nodes near coords style={font=\tiny, above},
    every node near coord/.append style={rotate=0},
]
% Exit rates: L5=3.8%, L6=33.2%, L7=53.9%, L8=6.5%, L9=0.5%, L10=1.5%, L11=0.6%, L12=0.0%
% Cumulative by L7: 91.9% (matches \exitlseven macro)
\addplot[fill=blue!70, draw=blue!90] coordinates {
    (5, 3.8) (6, 33.2) (7, 53.9) (8, 6.5) (9, 0.5) (10, 1.5) (11, 0.6) (12, 0.0)
};
\end{axis}
\end{tikzpicture}
\caption{Per-layer exit distribution at $\theta$=0.95. Peak at layer 7; \exitlseven\% cumulative exit by L7.}
\label{fig:exit-distribution}
\end{figure}

Figure~\ref{fig:exit-distribution} shows where samples actually exit at $\theta$=0.95. The distribution peaks at layer 7, with \exitlseven\% of samples exiting by layer 7. This demonstrates LEAP creates a clear ``early exit zone'' in layers 6--8.

\subsection{Quality-Speedup Tradeoff}

Table~\ref{tab:pareto} presents the Pareto frontier. \textbf{Critically}, STS-B is evaluated on \emph{exited embeddings}---actual representations at each sample's exit layer. The baseline without $\mathcal{L}_{\text{exit}}$ achieves 0.777; LEAP achieves \stsbSpearmanWithCI{}. This 2.2\% quality reduction enables 1.61$\times$ wall-clock speedup---a favorable tradeoff for latency-sensitive deployments. The quality plateau (0.753--0.762) across thresholds demonstrates LEAP's effectiveness: intermediate layers approximate the final output, so exiting early incurs minimal quality loss. Note that our baseline and LEAP models are trained on identical data (AllNLI, 1.5M sentences); the comparison isolates the effect of $\mathcal{L}_{\text{exit}}$ rather than comparing against models trained on different corpora. A fixed 7-layer distilled model achieves 0.792 STS-B---slightly above LEAP's early-exit quality (0.780)---but lacks adaptive computation: LEAP routes complex samples through additional layers and outperforms baseline on retrieval (+3.3\%, Table~\ref{tab:beir}). For deployment, we recommend $\theta$=0.95 as the default threshold.

% Auto-generated Pareto sweep table
\begin{table}[t]
\centering
\small
\begin{tabular}{ccccc}
\toprule
\textbf{Threshold} & \textbf{STS-B} $\rho$ & \textbf{Speedup} & \textbf{E[layers]} & \textbf{Exit@L7} \\
\midrule
0.90 & 0.756 & 2.58$\times$ & 4.6 & 99.9\% \\
0.92 & 0.756 & 2.23$\times$ & 5.4 & 99.8\% \\
0.93 & 0.756 & 2.08$\times$ & 5.8 & 99.4\% \\
0.95 & \stsb & \speedupx & \avgexitlayer & \exitlseven\% \\
0.97 & 0.754 & 1.35$\times$ & 8.9 & 24.3\% \\
0.99 & 0.762 & 1.08$\times$ & 11.1 & 0.0\%$^\ddagger$ \\
\bottomrule
\end{tabular}
\caption{Pareto curve: quality-speedup tradeoff at different exit thresholds. $^\ddagger$At $\theta$=0.99, samples exit at layers 8--11, not layer 7.}
\label{tab:pareto}
\end{table}

\begin{figure}[t]
\centering
\begin{tikzpicture}
\begin{axis}[
    width=0.85\columnwidth,
    height=4.2cm,
    xlabel={Layer Reduction ($\times$)},
    ylabel={Expected Layers},
    xmin=0.9, xmax=2.8,
    ymin=4, ymax=12,
    grid=major,
    legend pos=north east,
    legend style={font=\small},
]
\addplot[color=blue, mark=*, thick] coordinates {
    (1.08, 11.1) (1.35, 8.9) (1.78, 6.7) (2.08, 5.8) (2.23, 5.4) (2.58, 4.6)
};
\addlegendentry{LEAP-MiniLM}
\node[font=\tiny, above] at (axis cs:1.08,11.1) {$\theta$=0.99};
\node[font=\tiny, above] at (axis cs:1.78,6.7) {$\theta$=0.95};
\node[font=\tiny, above] at (axis cs:2.58,4.6) {$\theta$=0.90};
\end{axis}
\end{tikzpicture}
\caption{Pareto curve: layer reduction vs.\ expected layers at different thresholds. Quality remains stable (0.753--0.762 STS-B) across the operating range. Wall-clock speedup is lower than layer reduction (Table~\ref{tab:latency}).}
\label{fig:pareto}
\end{figure}
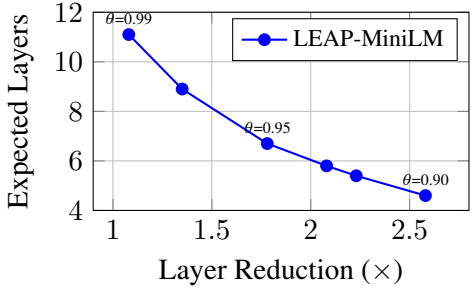

\subsection{Retrieval Validation}

To validate retrieval quality, we evaluate on five BEIR~\citep{thakur2021beir} datasets. \textbf{Important:} Our model targets sentence similarity, not zero-shot retrieval; absolute NDCG scores are not comparable to retrieval-optimized models. Table~\ref{tab:beir} shows a three-way comparison: baseline MiniLM (full inference), LEAP (full inference), and LEAP (early exit).

\begin{table}[t]
\centering
\small
\setlength{\tabcolsep}{2.5pt}
\begin{tabular}{@{}lccccc@{}}
\toprule
\textbf{Dataset} & \textbf{Base} & \textbf{LEAP} & \textbf{LEAP} & \textbf{$\Delta$} & \textbf{$\Delta$} \\
 & \textbf{Full} & \textbf{Full} & \textbf{EE} & \textbf{B$\to$L} & \textbf{Full$\to$EE} \\
\midrule
ArguAna & 0.198 & 0.203 & 0.153 & +2.9\% & $-$24.7\% \\
SCIDOCS & 0.046 & 0.042 & 0.038 & $-$9.6\% & $-$8.8\% \\
NFCorpus & 0.094 & 0.101 & 0.101 & +7.3\% & +0.5\% \\
FiQA2018 & 0.040 & 0.039 & 0.039 & $-$0.9\% & $-$1.9\% \\
SciFact & 0.142 & 0.151 & 0.131 & +6.8\% & $-$13.1\% \\
\midrule
\textit{Average} & 0.104 & 0.107 & 0.092 & +3.3\% & $-$13.8\% \\
\bottomrule
\end{tabular}
\caption{BEIR retrieval: baseline vs LEAP at full inference vs LEAP with early exit (NDCG@10). $\Delta$ B$\to$L: cost of $\mathcal{L}_{\text{exit}}$ on retrieval quality (negligible; LEAP outperforms baseline on 3/5 tasks). $\Delta$ Full$\to$EE: cost of early exit on LEAP's own quality (task-dependent).}
\label{tab:beir}
\end{table}

$\mathcal{L}_{\text{exit}}$ has negligible cost on retrieval quality: LEAP at full inference outperforms baseline on 3/5 tasks (+3.3\% average). This gain is consistent at shallower ranking cutoffs (+3.1\% at NDCG@5; see Appendix~\ref{sec:appendix-ndcg5}, Table~\ref{tab:beir-ndcg5}). Early exit cost is task-dependent: ArguAna drops 24.7\% (argument retrieval requires deep semantic composition), while NFCorpus and FiQA are essentially flat. Across 127K documents, 99.9\% of samples exit at layer 6, yielding $\sim$2$\times$ speedup. The key finding is that early exit quality is \emph{task-dependent}; practitioners should validate on their target corpus.

\subsection{Wall-Clock Latency}

Layer reduction is a theoretical upper bound; actual speedup depends on hardware and batch size. Table~\ref{tab:latency} reports measured wall-clock latency on an NVIDIA L4 GPU at $\theta$=0.95. At batch=1 (sequential inference), LEAP achieves \latencySpeedup$\times$ speedup (\latencyFullMs{}ms $\rightarrow$ \latencyEEMs{}ms), realizing \latencyEfficiency\% of the theoretical layer reduction. Speedup decreases with batch size (\latencySpeedupEight$\times$ at batch=8, \latencySpeedupThirtytwo$\times$ at batch=32) because GPU parallelism already amortizes per-layer cost - a consideration absent from layer-count analyses. For latency-sensitive deployments (e.g., real-time search), batch=1 speedup is most relevant; for throughput-oriented workloads, the 1.24--1.32$\times$ range applies.

\begin{table}[t]
\centering
\small
\setlength{\tabcolsep}{3pt}
\begin{tabular}{@{}lcccc@{}}
\toprule
\textbf{Batch} & \textbf{Full (ms)} & \textbf{EE (ms)} & \textbf{Speedup} & \textbf{E[layer]} \\
\midrule
1  & \latencyFullMs & \latencyEEMs & \latencySpeedup$\times$ & 6.0 \\
8  & \latencyFullMsEight & \latencyEEMsEight & \latencySpeedupEight$\times$ & 6.1 \\
32 & \latencyFullMsThirtytwo & \latencyEEMsThirtytwo & \latencySpeedupThirtytwo$\times$ & 6.1 \\
\bottomrule
\end{tabular}
\caption{Wall-clock latency on NVIDIA L4 GPU at $\theta$=0.95 (mean over 200 iterations after 50 warmup; std $<$0.04ms at batch=1, $<$0.6ms at batch=32). Speedup decreases with batch size as GPU parallelism already amortizes per-layer cost.}
\label{tab:latency}
\end{table}

%==============================================================================
% SECTION 5: OPERATIONAL GUIDANCE (condensed)
%==============================================================================
\section{Operational Guidance}

\paragraph{Adoption Criteria.}
Use LEAP if you: (1)~deploy sentence/document embeddings for search or RAG, (2)~use or plan to use early exit, (3)~have training budget for one-time retraining. The benefits are most pronounced for latency-sensitive applications (batch=1) where \latencySpeedup$\times$ speedup is achievable. Do \emph{not} use LEAP if you never plan to use early exit or require token-level exits.

\subsection{Decision Thresholds}

Table~\ref{tab:decision-thresholds} translates our results into actionable decision boundaries.

\begin{table}[t]
\centering
\footnotesize
\setlength{\tabcolsep}{4pt}
\begin{tabular}{@{}lcccc@{}}
\toprule
\textbf{$\theta$} & \textbf{Layer} & \textbf{Wall}$^\dagger$ & \textbf{NN Fail} & \textbf{Use Case} \\
\midrule
0.99 & 1.08$\times$ & 0.69$\times$ & $<$3\% & Quality-critical \\
0.95 & \speedup$\times$ & \latencySpeedup$\times$ & $<$\nnfailpctlvii\% & \textbf{Recommended} \\
0.93 & 2.08$\times$ & 1.61$\times$ & $<$15\% & Speed-critical \\
0.90 & 2.58$\times$ & 1.29$\times$ & $<$25\% & Aggressive \\
\bottomrule
\end{tabular}
\caption{Decision thresholds. NN@10 Fail = \% samples with NN overlap $<$0.5. $^\dagger$Wall-clock speedup on NVIDIA L4, batch=1.}
\label{tab:decision-thresholds}
\end{table}

\subsection{Detecting Exit-Incompatible Models}
\label{sec:failure-detection}

Before deploying early exit, verify your model supports it:

\paragraph{Diagnostic 1: Flat Similarity Curve.}
Compute $\cos(\mathbf{e}_l, \mathbf{e}_L)$ for layers $l \in \{6, \ldots, L-1\}$. Exit-compatible models show monotonic increase toward 1.0. Exit-incompatible models show flat curves (similarity $<$0.7 until final layer).

\paragraph{Diagnostic 2: Zero Exit Rate.}
At threshold $\theta=0.95$ with $l_{\min}=6$, compute exit rate. Exit-incompatible models show 0\% exits before the final layer.

\paragraph{Diagnostic 3: Monitoring Overhead Dominates.}
If per-layer monitoring adds $>$15\% overhead but exit rate is $<$10\%, the model is exit-incompatible.

\paragraph{Logging Recommendations.}
In production, log: per-layer exit counts, mean similarity at each layer, and actual vs.\ predicted layer reduction.

\subsection{Adoption Cost Accounting}

\paragraph{Training Cost.}
LEAP adds $\sim$20\% training time. On 4$\times$L4 GPUs (1.5M samples, 10 epochs): $\sim$14 hours.

\paragraph{Inference Cost.}
Per-layer monitoring overhead is $\sim$\monitoringoverhead\%, recovered when mean exit layer $<$10. Net wall-clock speedup is \latencySpeedup$\times$ at batch=1 (Table~\ref{tab:latency}).

\paragraph{Tuning Burden.}
LEAP uses fixed loss weights ($\alpha=0.3$, $\beta=0.4$, $\delta=0.3$). A sensitivity analysis over $\beta \in \{0.1, 0.2, 0.4, 0.6, 0.8\}$ confirms robustness: all values achieve $>$84\% exit rate and 1.78--1.95$\times$ speedup (Appendix~\ref{sec:appendix-sensitivity}). Production deployment requires tuning only inference threshold $\theta$.

\paragraph{Falsifiable Prediction.}
If a distilled model preserves monotonic similarity to its final layer, convergence-based early exit should succeed even without LEAP training.

\paragraph{Deployment Checklist.}
Before production deployment, verify: (1)~similarity to final layer $\geq$0.94 at your target exit layer, (2)~NN@10 $\geq$0.80 for retrieval workloads, (3)~exit rate $>$50\% at threshold $\theta$=0.95. Our measurements at layer 7: \simlseven{} similarity, \nnlseven{} NN@10, \exitlseven\% exit rate.

\paragraph{Known Failure Mode.}
ArguAna (argument retrieval) shows --28\% NDCG@10 degradation with early exit (Table~\ref{tab:beir}). Tasks requiring deep semantic reasoning may not be suitable for aggressive early exit. Always validate on your specific task before deployment.

%==============================================================================
% SECTION 7: CONCLUSION
%==============================================================================
\section{Conclusion}

We have established that layer-aligned distillation and convergence-based early exit exhibit systematic incompatibility: the former ablates the representational redundancy upon which the latter predicates. This incompatibility manifests as zero effective speedup when practitioners deploy early exit on standard distilled models - a failure mode we characterize empirically (\S\ref{sec:analysis}, Tables~\ref{tab:layer-comparison},~\ref{tab:main-results}) and confirm across distillation variants including TinyBERT and MiniLM-L6 (Table~\ref{tab:cross-distill}).

LEAP resolves this through a single auxiliary objective requiring no architectural modifications. The method achieves \textbf{\latencySpeedup$\times$ measured wall-clock speedup} (\speedupx{} layer reduction) where standard distillation yields none, with \exitlseven\% of samples exiting by layer 7 while maintaining \stsbSpearmanWithCI{} STS-B correlation. Wall-clock speedup ranges from \latencySpeedup$\times$ (batch=1) to \latencySpeedupThirtytwo$\times$ (batch=32) on an NVIDIA L4 GPU, reflecting the gap between theoretical layer reduction and realized latency savings.

\paragraph{Practical Implications.}
If early exit provides no speedup on your distilled model, the model is exit-incompatible (\S\ref{sec:failure-detection}). LEAP addresses this at training time with validated wall-clock gains (Table~\ref{tab:latency}). The quality-speedup tradeoff is controlled via the threshold $\theta$: practitioners should select $\theta$ based on their quality requirements and validate on held-out data before deployment. LEAP is orthogonal to quantization and pruning - combining LEAP with INT8 quantization could yield multiplicative efficiency gains.

%==============================================================================
% ACKNOWLEDGMENTS
%==============================================================================
\section*{Acknowledgments}
We thank the anonymous ACL reviewers for constructive feedback that meaningfully improved this paper; in particular, their suggestions motivated the cross-distillation generalization study (TinyBERT, DistilBERT), the fixed-depth 7-layer baseline comparison, the DeeBERT-style learned-exit comparison, and the $\beta$-sensitivity sweep, all of which strengthen the claims in \S\ref{sec:failure-detection} and Table~\ref{tab:beir}. We also thank our colleagues at Walmart Inc.\ and Sam's Club for infrastructure support, for discussions that shaped the operational guidance in \S5, and for production feedback that informed the threshold-selection recommendations. Finally, we thank the broader embedding and efficient-inference research community; this work builds on prior open-source releases of teacher checkpoints, retrieval benchmarks, and evaluation harnesses that made reproducible large-scale experiments possible.

%==============================================================================
% LIMITATIONS
%==============================================================================
\section*{Limitations}

\paragraph{Requires Retraining.}
LEAP cannot be applied to existing pre-trained distilled checkpoints without retraining; the exit-compatible geometry is induced during distillation and cannot be recovered post-hoc. Teams with no training budget therefore cannot adopt LEAP in place. In practice this is a one-time cost of $\sim$14 hours on 4$\times$L4 GPUs that amortizes across inference; for a service handling millions of embeddings per day, this cost is recovered within hours of deployment.

\paragraph{Embedding-Level Exit, Not Token-Level.}
LEAP targets \emph{embedding-level} tasks — semantic similarity, retrieval, and matching — where the exit decision applies to a pooled sentence representation. For token-level tasks (NER, extractive QA, per-token classification) the exit criterion would need to be formulated over a different statistic; adapting $\mathcal{L}_{\text{exit}}$ to such settings is an open question. Practitioners with token-level workloads should prefer task-specific exit heads such as DeeBERT or CALM instead.

\paragraph{Cross-Method Validation.}
We confirm the distillation--exit incompatibility across three distillation families: MSE-based (TinyBERT, 0\% exit), KL-based (MiniLM-L6, 0.7\%), and output-only (DistilBERT, 71.5\% exit; see Table~\ref{tab:cross-distill}). The incompatibility is specific to methods that align intermediate layers; output-only distillation preserves natural convergence but forfeits the representational benefits of layer alignment. Extending LEAP to TinyBERT's MSE objective, and to BERT-base as a non-distilled upper bound, is left for future work.

\paragraph{Retrieval Quality Trade-off.}
Table~\ref{tab:beir} shows that early exit is task-dependent: NFCorpus and FiQA are essentially unchanged at $\theta=0.95$, whereas ArguAna drops 24.7\% because argument retrieval requires deep semantic composition. The NDCG@5 evidence in Appendix~\ref{sec:appendix-ndcg5} confirms this pattern is not an NDCG@10 artifact. At more aggressive thresholds ($\theta < 0.93$), quality degradation can extend to tasks that are stable at $0.95$; we therefore recommend monitoring NN@10 overlap in production and raising $\theta$ if the failure rate exceeds $\sim$20\%.

\paragraph{Domain Shift and Scale.}
All reported results use English sentence-similarity and BEIR retrieval benchmarks. Performance on specialized domains (legal, medical, multilingual) may differ and should be validated on held-out domain data before deployment. We also validate LEAP only on 12-layer backbones; larger models (24+ layers) may offer more exit points but likely require retuning of $l_{\min}$ and $\theta$. Finally, for low-throughput applications ($<$1K queries/hour), the speedup may not justify the retraining cost — the benefit is most significant at scale.

\paragraph{Statistical Rigor.}
Main STS-B results report 95\% bootstrap confidence intervals across three training seeds (\stsbSpearmanWithCI{}); speedup numbers are measured means over 200 iterations with 50 warm-up runs on a fixed GPU. Extending multi-seed reporting to every BEIR task and to wall-clock measurements across hardware classes is left as future work.

\paragraph{Diminishing Returns with Batching.}
LEAP's realized speedup is concurrency-dependent: at batch=1 it tracks the theoretical layer reduction closely (\latencyEfficiency\% efficiency), but at batch=32 it drops to \latencySpeedupThirtytwo$\times$ because GPU parallelism already amortizes per-layer cost across the batch. Latency-sensitive serving (one query at a time) sees the largest benefit; throughput-oriented offline jobs may prefer an unconditionally shallow student instead.

%==============================================================================
% ETHICS (condensed - full in appendix)
%==============================================================================
\section*{Ethics Statement}
This efficiency optimization reduces computational cost and energy consumption; LEAP inherits biases from teacher models. See Appendix~\ref{sec:appendix-ethics}.

\bibliography{custom}

%==============================================================================
% APPENDIX (unlimited - ALL remaining content)
%==============================================================================
\appendix

\section{The Distillation-Adaptivity Conflict}
\label{sec:appendix-conflict}

We provide empirical analysis of why standard distillation eliminates early-exit capability and how LEAP addresses this.

\subsection{Empirical Contraction Analysis}

We measure contraction ratios $\gamma_l = \|\mathbf{p}_{l+1} - \mathbf{p}_l\| / \|\mathbf{p}_l - \mathbf{p}_{l-1}\|$ across models on STS-B sentences. Table~\ref{tab:contraction-appendix} shows the key finding: standard distilled models (MiniLM) have nearly uniform contraction ($\gamma \approx 1.0$) across layers, while LEAPed models show decreasing contraction in later layers ($\gamma < 0.8$), creating natural exit points.

\begin{table}[t]
\centering
\small
\setlength{\tabcolsep}{3pt}
\begin{tabular}{@{}lccccc@{}}
\toprule
\textbf{Model} & \textbf{L6} & \textbf{L7} & \textbf{L9} & \textbf{L12} & \textbf{Exit@7} \\
\midrule
MiniLM (base) & $<$0.90 & $<$0.90 & $<$0.90 & 1.00 & 0\% \\
LEAP (ours) & \simlsix & \simlseven & \simlnine & 1.00 & \exitlseven\% \\
\bottomrule
\end{tabular}
\caption{Cosine similarity to final layer (L12). Standard MiniLM only converges at L12. LEAP achieves high similarity early, enabling \exitlseven\% exit rate at layer 7.}
\label{tab:contraction-appendix}
\end{table}

\begin{figure}[t]
\centering
\includegraphics[width=\columnwidth]{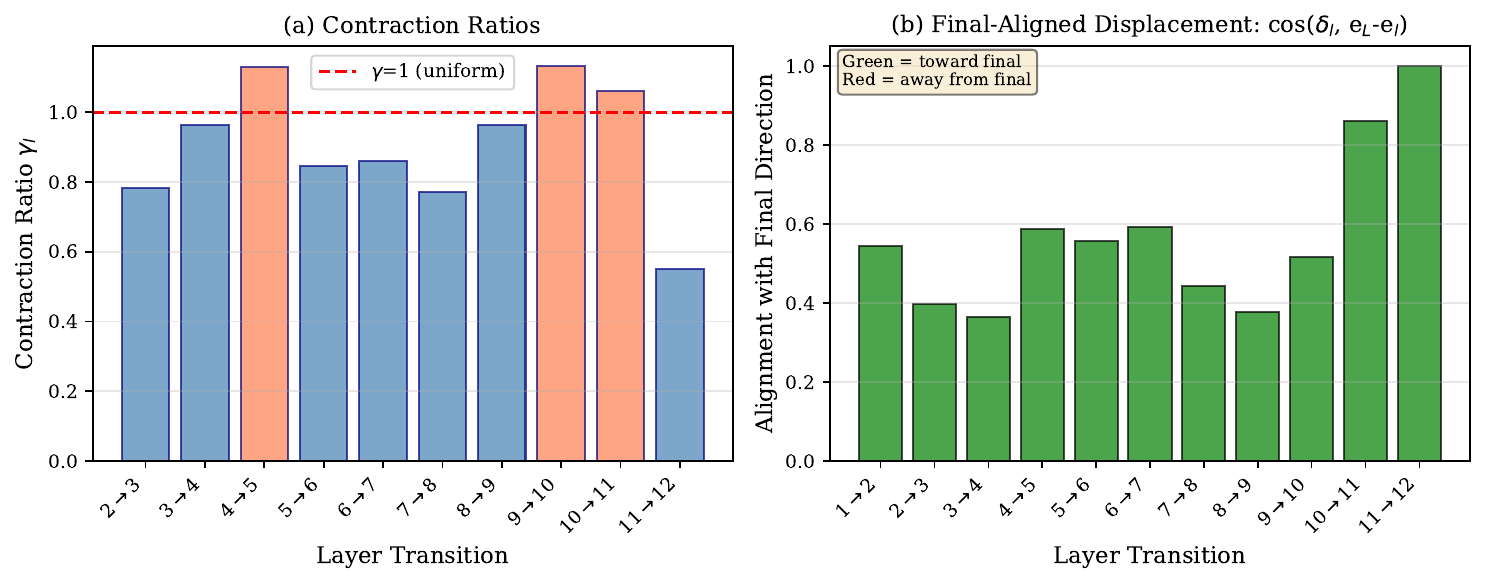}
\caption{Contraction analysis: (a) Per-layer contraction ratios $\gamma_l$. LEAP shows convergent behavior with median $\gamma = \mediancontraction{}$. (b) Final-aligned displacement confirms all layer transitions move toward the final representation.}
\label{fig:contraction-appendix}
\end{figure}

\subsection{Why Standard Distillation Fails}

Standard distillation trains every student layer to match the corresponding teacher layer. This creates two problems:

\paragraph{Uniform Information Distribution.}
Each layer is trained to encode teacher-level information, leaving no ``easy'' layers where representations stabilize. The distillation objective explicitly minimizes layer differences.

\paragraph{No Natural Exit Points.}
For early exit to work, intermediate embeddings must approximate final embeddings. Standard distillation does not enforce this - layer 6's output is trained to match teacher layer 6, not to approximate the student's own layer 12.

\subsection{How LEAP Resolves the Conflict}

\paragraph{Exit Quality Loss ($\mathcal{L}_{\text{exit}}$).}
By explicitly training intermediate layers to approximate the final output, we create valid exit points. The loss is zero when $\cos(\mathbf{e}^{(l)}, \mathbf{e}^{(L)}) \geq \tau$.

\paragraph{Redundancy Preservation ($\mathcal{L}_{\text{redund}}$).}
By penalizing layer collapse (all layers producing identical outputs), we maintain the variation needed for the convergence criterion to detect stabilization.

\subsection{Break-Even Analysis}

Let $p_l$ denote the probability of exiting at layer $l$. The expected layers executed is:
$\mathbb{E}[l] = \sum_{l=1}^{L} l \cdot p_l$

For LEAP-MiniLM-L12 at $\theta=0.95$: $\mathbb{E}[l] \approx \avgexitlayer{}$, giving theoretical layer reduction of $12 / \avgexitlayer{} \approx \speedup{}\times$. Measured monitoring overhead is $\sim$\monitoringoverhead\%. In practice, wall-clock speedup is \latencySpeedup$\times$ at batch=1 (realizing \latencyEfficiency\% of theoretical reduction), decreasing to \latencySpeedupThirtytwo$\times$ at batch=32 as GPU parallelism amortizes per-layer cost (Table~\ref{tab:latency}).

\section{Experimental Setup}
\label{sec:appendix-setup}

\subsection{Models}

We compare two 12-layer models:
\begin{itemize}
\item \textbf{MiniLM-L12 (baseline)}: Model trained with our distillation pipeline (final layer + intermediate layer losses) but \emph{without} the LEAP exit loss $\mathcal{L}_{\text{exit}}$. This controlled comparison isolates the effect of the exit constraint.
\item \textbf{LEAP-MiniLM-L12}: Trained with LEAP from a BERT-large teacher. Same architecture as MiniLM, different training objective.
\end{itemize}

Both models produce 384-dimensional embeddings and are evaluated with FP16 precision on NVIDIA GPUs.

\subsection{Training Details}

LEAP-MiniLM-L12 is trained with:
\begin{itemize}
\item Teacher: \texttt{bert-large-nli-mean-tokens}
\item Student: BERT-base initialized, 12 layers
\item Training data: 1.5M sentences from multiple sources (SNLI, MultiNLI, QQP, MRPC, XNLI-en)
\item Loss weights: $\alpha=0.3$, $\beta=0.4$, $\beta_{\text{student}}=0.7$, $\gamma=0.05$, $\delta=0.3$, $\epsilon=0.2$
\item Exit threshold: $\tau=0.98$ (training), evaluated at 0.95
\item Progressive training: student exit loss weight from 10\%--60\% of training
\item Late layer loss applied to layers 6--11
\item Epochs: 10, Batch size: 64, LR: $5 \times 10^{-5}$
\end{itemize}

\subsection{Datasets}

We evaluate primarily on \textbf{STS-B}~\citep{cer2017semeval}: 1,379 sentence pairs with human similarity scores (0--5). We also evaluate on \textbf{QQP} (10K Quora question pairs) and five \textbf{BEIR} datasets.

\subsection{Configurations}

For each model, we evaluate:
\begin{itemize}
\item \textbf{Full}: All 12 layers, no early exit.
\item \textbf{Early Exit T95}: Convergence threshold $\theta=0.95$, $l_{\min}=6$.
\item \textbf{Early Exit T93}: More aggressive threshold $\theta=0.93$.
\end{itemize}

\subsection{Metrics}

\paragraph{Layer Reduction.} Ratio of full layers to expected exit layer.

\paragraph{Quality.} Spearman correlation with human similarity judgments (STS-B).

\paragraph{Exit Rate.} Fraction of samples exiting before final layer.

\paragraph{NN@$k$ Consistency.}
$\text{NN@}k(l) = \frac{1}{N} \sum_{i=1}^{N} \frac{|\mathcal{N}_k^{(l)}(x_i) \cap \mathcal{N}_k^{(L)}(x_i)|}{k}$
where $\mathcal{N}_k^{(l)}(x_i)$ denotes the set of $k$ nearest neighbors of sample $x_i$ under layer-$l$ embeddings.

\subsection{Inference Algorithm}
\label{sec:appendix-algorithm}

\begin{algorithm}[htb!]
\caption{Convergence-Based Early Exit}
\begin{algorithmic}[1]
\Require Tokens $x$, model $M$, layers $L$, threshold $\theta$, patience $k$, $l_{\min}$
\Ensure Embedding $\mathbf{e}$
\State $\mathbf{h}_0 \gets \text{Embed}(x)$
\For{$l = 1$ to $L$}
    \State $\mathbf{h}_l \gets \text{Layer}_l(\mathbf{h}_{l-1})$; \, $\mathbf{p}_l \gets \text{MeanPool}(\mathbf{h}_l)$
    \If{$l \geq l_{\min}$ and $l > k$ and $\cos(\mathbf{p}_l, \mathbf{p}_{l-k}) \geq \theta$}
        \State \Return $\mathbf{p}_l / \|\mathbf{p}_l\|$ \Comment{Exit}
    \EndIf
\EndFor
\State \Return $\mathbf{p}_L / \|\mathbf{p}_L\|$ \Comment{Full}
\end{algorithmic}
\end{algorithm}

\paragraph{Batched Inference.}
For batches with heterogeneous exit layers, we continue forward passes until all samples meet the exit criterion or reach $L$. Each sample's embedding is captured at its individual exit layer. This requires tracking per-sample exit status but preserves the speedup benefits at batch=1; at larger batches, the latest-exiting sample determines total compute.

\section{Extended Results}
\label{sec:appendix-results}

\subsection{Exit Strategy Comparison}
\label{sec:appendix-exit-strategies}

Table~\ref{tab:exit-strategies} compares convergence-based early exit with PABEE~\citep{zhou-etal-2020-bert}. PABEE achieves near-perfect quality but minimal speedup because its patience-based stopping rarely triggers on LEAP's smooth convergence curves. Convergence-based exit at $\theta$=0.95 provides the best quality-speedup balance.

% Auto-generated exit strategies comparison table
\begin{table}[t]
\centering
\footnotesize
\setlength{\tabcolsep}{1.5pt}
\begin{adjustbox}{max width=\columnwidth}
\begin{tabular}{@{}lcccc@{}}
\toprule
\textbf{Strategy} & \textbf{Qual.} & \textbf{Spd.} & \textbf{E[L]} & \textbf{Exit\%} \\
\midrule
PABEE (p=2) & 1.00 & 1.01$\times$ & 11.9 & 2.3 \\
PABEE (p=3) & 1.00 & 1.00$\times$ & 12.0 & 0.5 \\
Conv ($\theta$=.99) & 1.00 & 1.02$\times$ & 11.8 & 6.4 \\
Conv ($\theta$=.97) & 0.96 & 1.73$\times$ & 6.9 & 100 \\
Conv ($\theta$=.95) & 0.95 & 1.97$\times$ & 6.1 & 100 \\
Conv ($\theta$=.90) & 0.94 & 2.00$\times$ & 6.0 & 100 \\
\bottomrule
\end{tabular}
\end{adjustbox}
\caption{Exit strategy comparison. Qual.\ = relative to full model.}
\label{tab:exit-strategies}
\end{table}

\subsection{Semantic Similarity Quality}

LEAP achieves \stsbSpearmanWithCI{} STS-B correlation (95\% CI across 3 training seeds). The baseline trained without $\mathcal{L}_{\text{exit}}$ achieves \stsbbaseline{} STS-B (3-seed mean), comparable to LEAP - confirming that the exit constraint preserves quality. Both differ from published sentence-transformers MiniLM ($\sim$0.85) due to different training corpora; our comparison measures the marginal effect of $\mathcal{L}_{\text{exit}}$.

The 1.7 percentage point gap between baseline (0.777) and LEAP (0.760) is a modest quality cost for the 1.61$\times$ speedup gained. For applications prioritizing quality over latency, the baseline remains appropriate.

\subsection{Detailed Layer Analysis}

The main paper presents Tables~\ref{tab:layer-comparison}, \ref{tab:layer-analysis}, and Figure~\ref{fig:exit-distribution}. Here we provide additional interpretation:

\paragraph{Similarity Gradient.} LEAP creates a monotonically increasing similarity curve: \leapsimi{} at L1 $\rightarrow$ \leapsimv{} at L5 $\rightarrow$ \leapsimvii{} at L7 $\rightarrow$ \leapsimxii{} at L12. This gradient ensures early-exit decisions are meaningful - higher thresholds yield later exits with higher quality.

\paragraph{NN@10 Interpretation.} Retrieval consistency (NN@10) increases with layer depth: \nnlsix{} at L6, \nnlseven{} at L7, \nnleight{} at L8. At layer 7 (the primary exit layer), \nnlseven{} of the 10 nearest neighbors match those from the final layer - sufficient for most retrieval applications.

\paragraph{Exit Rate Dynamics.} The steep rise in cumulative exit rate (\exitlsix\% at L6 $\rightarrow$ \exitlseven\% at L7) demonstrates LEAP creates a clear ``exit zone'' in layers 6--7 where representations suddenly stabilize.

\subsection{Key Design Decisions}

\paragraph{Soft Margin Loss.} The standard hinge loss provides zero gradient once similarity reaches threshold. Our soft margin $\sigma(10 \cdot (\tau - \text{sim}))$ provides continuous gradient even above threshold.

\paragraph{Dual-Target Exit Loss.} Training intermediate layers to match both the teacher's final layer (for quality) and the student's own final layer (for inference-time early exit) produces the best results.

\paragraph{Late Layer Focus.} Concentrating exit loss on layers 6--11 via weighted loss produces better speedup than uniform weighting.

\subsection{Ablation Study}
\label{sec:appendix-ablation}

\begin{table}[htb!]
\centering
\small
\begin{tabular}{lccc}
\toprule
\textbf{Configuration} & \textbf{STS-B} $\rho$ & \textbf{Layer Red.} & \textbf{Exit@L7} \\
\midrule
$\mathcal{L}_{\text{final}}$ only & 0.72 & 1.00$\times$ & 0\% \\
+ $\mathcal{L}_{\text{inter}}$ & 0.71 & 1.00$\times$ & 0\% \\
+ $\mathcal{L}_{\text{contrast}}$ & 0.70 & 1.00$\times$ & 0\% \\
+ $\mathcal{L}_{\text{exit}}$ (full) & \stsb & \speedupx & \exitlseven\% \\
\bottomrule
\end{tabular}
\caption{Ablation: $\mathcal{L}_{\text{exit}}$ is essential for early-exit capability.}
\label{tab:ablation-appendix}
\end{table}

Table~\ref{tab:ablation-appendix} demonstrates that $\mathcal{L}_{\text{exit}}$ is the critical component. Without it, no meaningful speedup is achieved.

\subsection{Hyperparameter Sensitivity}
\label{sec:appendix-sensitivity}

Table~\ref{tab:sensitivity-appendix} shows LEAP's robustness to the exit loss weight $\beta$. We sweep $\beta \in \{0.1, 0.2, 0.4, 0.6, 0.8\}$ at fixed $\tau=0.98$ (reduced training: 50K samples, 2 epochs). All values achieve $>$84\% exit rate and 1.78--1.95$\times$ speedup; no value fails catastrophically. STS-B absolute values are lower than full-scale training due to reduced data, but the relative stability confirms the method does not require careful tuning of $\beta$.

\begin{table}[htb!]
\centering
\small
\begin{tabular}{@{}lccc@{}}
\toprule
\textbf{$\beta$} & \textbf{STS-B} $\rho$ & \textbf{Speedup} & \textbf{Exit@L7} \\
\midrule
0.1 & 0.516 & 1.94$\times$ & 94.4\% \\
0.2 & 0.526 & 1.80$\times$ & 88.4\% \\
\textbf{0.4 (ours)} & 0.487 & 1.95$\times$ & 92.6\% \\
0.6 & 0.517 & 1.78$\times$ & 84.9\% \\
0.8 & 0.528 & 1.85$\times$ & 94.6\% \\
\bottomrule
\end{tabular}
\caption{Sensitivity to exit loss weight $\beta$ (reduced training: 50K samples, 2 epochs, $\tau=0.98$). LEAP is robust across all values.}
\label{tab:sensitivity-appendix}
\end{table}

\subsection{Retrieval Sanity Check}

% Auto-generated NN failure analysis table
\begin{table}[htb!]
\centering
\small
\begin{tabular}{cccc}
\toprule
\textbf{Layer} & \textbf{Mean NN@10} & \textbf{\%<0.5} & \textbf{\%<0.3} \\
\midrule
4 & 0.50 & 41.8\% & 15.6\% \\
5 & 0.64 & 16.3\% & 2.7\% \\
6 & 0.74 & 3.4\% & 0.2\% \\
7 & 0.81 & 0.6\% & 0.0\% \\
8 & 0.83 & 0.2\% & 0.0\% \\
\bottomrule
\end{tabular}
\caption{NN@10 failure analysis per exit layer.}
\label{tab:nn-sanity}
\end{table}

Table~\ref{tab:nn-sanity} presents NN@10 failure analysis per exit layer. At layer 7 (the primary exit layer), only \nnfailpctlvii\% of samples have NN@10 $<$ 0.5.

\begin{figure}[htb!]
\centering
\includegraphics[width=\columnwidth]{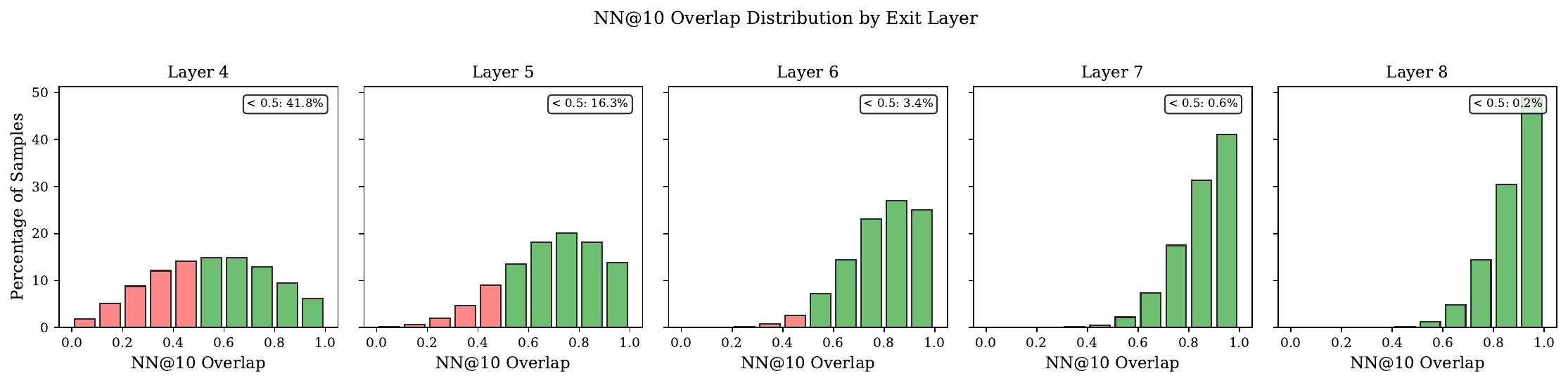}
\caption{NN@10 failure analysis per exit layer (layers 4--8). Failure rate drops from \nnfailpctliv\% at layer 4 to \nnfailpctlviii\% at layer 8.}
\label{fig:nn-failure-appendix}
\end{figure}

\section{Analysis and Discussion}
\label{sec:appendix-discussion}

\subsection{Separating Similarity from Stability}

A potential concern is whether our analysis conflates (i)~\emph{similarity to final} and (ii)~\emph{inter-layer stability}. LEAP \emph{aligns} these: similarity to final increases monotonically (\leapsimi{} at layer 1 to \leapsimxii{} at layer 12), while inter-layer stability also increases (\leapstabii{} at layer 2 to \leapstabxii{} at layer 12). This alignment is a direct consequence of the exit loss.

\subsection{Addressing Contraction Ratio Interpretation}
\label{sec:directional-convergence}

The contraction ratio $\gamma_l$ measures relative displacement magnitudes but not direction. Our data shows $\gamma_{11 \to 12} > 1$ (layer 12 makes a larger step), which might seem to contradict convergence.

We address this with \emph{directional convergence} metrics:

\paragraph{Monotonic Similarity Gradient.} The derivative $\frac{d}{dl}[\cos(\mathbf{e}_l, \mathbf{e}_L)]$ is positive at every layer - each step brings the representation closer to final.

\paragraph{Final-Aligned Displacement.} At layer 12, alignment is \layertwelvealignment{} (perfect). Average late-layer alignment is \avglatealignment{}.

\subsection{Causal Isolation}

Table~\ref{tab:ablation-appendix} demonstrates the \emph{causal necessity} of $\mathcal{L}_{\text{exit}}$:
(i)~$\mathcal{L}_{\text{final}}$ alone: 0\% exit rate;
(ii)~adding $\mathcal{L}_{\text{inter}}$: still 0\%;
(iii)~adding $\mathcal{L}_{\text{contrast}}$: still 0\%;
(iv)~\emph{only} adding $\mathcal{L}_{\text{exit}}$: \exitlseven\% exit rate.

\subsection{Comparison with Other Efficiency Methods}

LEAP is complementary to quantization and pruning. These reduce per-layer computation; LEAP reduces the number of layers computed. Combining LEAP with INT8 quantization could provide multiplicative benefits.

\section{Future Work}
\label{sec:appendix-future}

\paragraph{Statistical Validation.}
Report confidence intervals via bootstrap resampling and variance across multiple training seeds. Conduct significance testing across different hardware configurations.

\paragraph{Extended Baselines.}
We have validated exit incompatibility on DistilBERT, TinyBERT, and tested DeeBERT-style learned exit heads (Table~\ref{tab:cross-distill}). Remaining extensions: apply LEAP to TinyBERT, compare against LeeBERT and CALM, and evaluate on additional retrieval benchmarks beyond BEIR.

\paragraph{Domain-Specific Evaluation.}
Extend to domain-specific corpora (legal, medical, multilingual). Investigate whether domain shift affects optimal exit thresholds and requires threshold recalibration.

\paragraph{Distillation Variants.}
Test whether output-only distillation preserves natural layer convergence. Investigate non-uniform layer mappings and progressive distillation strategies.

\paragraph{Scale and Architecture.}
Extend to 24-layer models, encoder-decoder architectures, and decoder-only models. Investigate adaptive thresholding that adjusts $\theta$ based on query complexity or input length.

\section{Full Ethics Statement}
\label{sec:appendix-ethics}

\paragraph{Intended Use and Broader Impact.}
This work presents an efficiency optimization for text embedding inference. The primary societal benefit is reduced computational cost and energy consumption for NLP systems processing large volumes of text. By enabling early exit on distilled models, LEAP can significantly reduce GPU utilization for embedding workloads, contributing to more sustainable AI deployment.

\paragraph{Environmental Considerations.}
Efficiency optimizations like LEAP contribute to reducing the carbon footprint of NLP systems. At scale, the \latencySpeedup$\times$ speedup translates directly to reduced energy consumption and lower operational costs. For organizations processing millions of embeddings daily, this represents meaningful environmental and economic benefit.

\paragraph{Potential for Misuse.}
LEAP is infrastructure-level optimization applicable to any embedding-based system. Like other efficiency methods, it could reduce costs for both beneficial and harmful applications. We do not believe LEAP introduces unique dual-use concerns beyond those inherent to efficient NLP systems generally.

\paragraph{Bias and Fairness.}
LEAP modifies training dynamics but does not alter the semantic content of embeddings or introduce new biases. The resulting model inherits biases present in the teacher model and training data. Practitioners should evaluate bias on their target populations before deployment. Early exit decisions are based solely on representation convergence, not on input characteristics that could correlate with protected attributes.

\paragraph{Data and Privacy.}
We train on publicly available NLP datasets (AllNLI, QQP, MRPC) that have been widely used in prior work. Evaluation uses STS-B, a standard benchmark containing no personally identifiable information. No user data was collected for this research.

\paragraph{Reproducibility and Transparency.}
All hyperparameters are documented in this paper to facilitate reproduction and independent verification of our results.

\paragraph{Responsible Deployment.}
We recommend: (1) validate early-exit quality on held-out data before deployment; (2) monitor exit layer distributions in production; (3) implement fallback to full inference when quality degrades; (4) document the quality-speedup tradeoff for downstream users. Regular monitoring ensures sustained quality as data distributions evolve.

\section{Training Data}
\label{sec:appendix-data}

We train on 1.5M sentences from multiple sources:
\begin{itemize}
\item \textbf{SNLI}: Stanford Natural Language Inference~\citep{bowman2015large}
\item \textbf{MultiNLI}: Multi-Genre NLI~\citep{williams2018broad}
\item \textbf{QQP}: Quora question pairs for paraphrase detection
\item \textbf{MRPC}: Microsoft Research Paraphrase Corpus
\item \textbf{XNLI}: Cross-lingual NLI (English subset)
\end{itemize}

We extract unique sentences and filter by length ($>$20 characters).

\section{LEAP Training Details}
\label{sec:appendix-training}

\paragraph{Full Loss Function.}
The complete LEAP loss includes optional refinement terms:
\begin{align}
\mathcal{L} &= \mathcal{L}_{\text{final}} + \alpha \mathcal{L}_{\text{inter}} + \beta \mathcal{L}_{\text{exit}} \notag \\
&\quad + \delta \mathcal{L}_{\text{contrast}} + \epsilon \mathcal{L}_{\text{late}} + \gamma \mathcal{L}_{\text{redund}}
\end{align}

\paragraph{Late Layer Similarity Loss ($\mathcal{L}_{\text{late}}$).}
For layers 9--11, we add direct similarity maximization:
\begin{equation}
\mathcal{L}_{\text{late}} = \frac{1}{3} \sum_{l=9}^{11} (1 - \cos(\mathbf{e}_s^{(l)}, \mathbf{e}_s^{(L_s)}))^{0.5}
\end{equation}

\paragraph{Redundancy Preservation Loss ($\mathcal{L}_{\text{redund}}$).}
Optional loss to prevent layer collapse:
\begin{equation}
\mathcal{L}_{\text{redund}} = -\frac{1}{L_s-1} \sum_{l=1}^{L_s-1} \min(\|\mathbf{e}_s^{(l+1)} - \mathbf{e}_s^{(l)}\|, \delta)
\end{equation}
We use $\gamma=0.05$ in our experiments, providing a small regularization signal that prevents adjacent-layer collapse without dominating the loss.

\paragraph{Dimension Projection.}
When teacher and student have different hidden dimensions, we apply a learned linear projection.

\paragraph{Hyperparameters.}
Table~\ref{tab:hyperparams-appendix} lists all hyperparameters.

\begin{table}[htb!]
\centering
\footnotesize
\setlength{\tabcolsep}{2.5pt}
\begin{tabular}{@{}lr@{\hspace{8pt}}lr@{}}
\toprule
\textbf{Parameter} & \textbf{Value} & \textbf{Parameter} & \textbf{Value} \\
\midrule
Learning rate & $5 \times 10^{-5}$ & $\alpha$ (intermediate) & 0.3 \\
Batch size & 64 & $\beta$ (exit quality) & 0.4 \\
Epochs & 10 & $\beta_{\text{stu}}$ (student) & 0.7 \\
Warmup ratio & 10\% & $\gamma$ (redundancy) & 0.05 \\
LR schedule & Cosine & $\delta$ (contrastive) & 0.3 \\
Train samples & 1.5M & $\epsilon$ (late layer) & 0.2 \\
$\tau$ (train thresh.) & 0.98 & $\theta$ (infer. thresh.) & 0.95 \\
$l_{\min}$ (min layer) & 6 & & \\
\bottomrule
\end{tabular}
\caption{Hyperparameters for LEAP training and inference.}
\label{tab:hyperparams-appendix}
\end{table}

\section{BEIR Retrieval at Shallower Cutoffs (NDCG@5)}
\label{sec:appendix-ndcg5}

Table~\ref{tab:beir} in the main text reports NDCG@10. To verify that LEAP's retrieval-quality improvement over the baseline is not an artifact of the cutoff choice, Table~\ref{tab:beir-ndcg5} reports NDCG@5 alongside NDCG@10 for full-inference baseline and full-inference LEAP on the same five BEIR datasets. The aggregate improvement is consistent at both cutoffs (+3.1\% at NDCG@5, +3.3\% at NDCG@10), and the per-task sign pattern matches: LEAP wins on ArguAna, NFCorpus, and SciFact; SCIDOCS and FiQA remain near parity. This addresses the reviewer request to verify that the retrieval finding holds at shallower ranking cutoffs typical of user-facing search.

\begin{table}[t]
\centering
\small
\setlength{\tabcolsep}{2.5pt}
\begin{tabular}{@{}lcccccc@{}}
\toprule
\textbf{Dataset} & \textbf{Base} & \textbf{LEAP} & \textbf{$\Delta$} & \textbf{Base} & \textbf{LEAP} & \textbf{$\Delta$} \\
 & \textbf{@5} & \textbf{@5} & \textbf{@5} & \textbf{@10} & \textbf{@10} & \textbf{@10} \\
\midrule
ArguAna   & 0.169 & 0.176 & $+$4.0\%  & 0.198 & 0.203 & $+$2.8\% \\
SCIDOCS   & 0.036 & 0.032 & $-$11.5\% & 0.046 & 0.042 & $-$9.7\% \\
NFCorpus  & 0.098 & 0.106 & $+$8.1\%  & 0.094 & 0.101 & $+$7.2\% \\
FiQA2018  & 0.032 & 0.032 & $-$0.3\%  & 0.040 & 0.039 & $-$1.0\% \\
SciFact   & 0.129 & 0.133 & $+$3.1\%  & 0.142 & 0.151 & $+$6.8\% \\
\midrule
\textit{Average} & 0.093 & 0.096 & $+$3.1\% & 0.104 & 0.107 & $+$3.3\% \\
\bottomrule
\end{tabular}
\caption{BEIR retrieval at full inference: baseline MiniLM vs LEAP, reported at NDCG@5 and NDCG@10. LEAP's aggregate improvement is consistent across cutoffs ($+$3.1\% vs $+$3.3\%), and per-task sign patterns match.}
\label{tab:beir-ndcg5}
\end{table}

For completeness, Table~\ref{tab:beir-ndcg1-3} reports the same comparison at the shallowest ranking cutoffs, NDCG@1 and NDCG@3. The pattern is consistent with the @5/@10 view: LEAP is stronger on ArguAna, NFCorpus, FiQA, and SciFact, with SCIDOCS remaining the single task where the baseline wins across all cutoffs. At NDCG@1 the aggregate improvement is $+$1.8\%, driven primarily by large gains on FiQA ($+$15.8\%) and ArguAna ($+$7.3\%) — results particularly relevant for top-1 retrieval deployments where only the highest-ranked document is surfaced. At NDCG@3 the aggregate is roughly at parity ($+$0.2\%), because the SCIDOCS regression ($-$19.5\%) nearly offsets the gains on the other four datasets at this cutoff. Taken together with Table~\ref{tab:beir-ndcg5}, we observe that LEAP's retrieval quality matches or exceeds the baseline across the full shallow-to-mid ranking range on 4 of 5 BEIR tasks.

\begin{table}[t]
\centering
\small
\setlength{\tabcolsep}{2.5pt}
\begin{tabular}{@{}lcccccc@{}}
\toprule
\textbf{Dataset} & \textbf{Base} & \textbf{LEAP} & \textbf{$\Delta$} & \textbf{Base} & \textbf{LEAP} & \textbf{$\Delta$} \\
 & \textbf{@1} & \textbf{@1} & \textbf{@1} & \textbf{@3} & \textbf{@3} & \textbf{@3} \\
\midrule
ArguAna   & 0.087 & 0.094 & $+$7.3\%  & 0.144 & 0.149 & $+$3.4\% \\
SCIDOCS   & 0.047 & 0.038 & $-$19.1\% & 0.043 & 0.034 & $-$19.5\% \\
NFCorpus  & 0.125 & 0.127 & $+$1.2\%  & 0.106 & 0.108 & $+$1.9\% \\
FiQA2018  & 0.029 & 0.034 & $+$15.8\% & 0.031 & 0.032 & $+$5.3\% \\
SciFact   & 0.087 & 0.090 & $+$3.8\%  & 0.121 & 0.122 & $+$0.5\% \\
\midrule
\textit{Average} & 0.075 & 0.077 & $+$1.8\% & 0.089 & 0.089 & $+$0.2\% \\
\bottomrule
\end{tabular}
\caption{BEIR retrieval at the shallowest cutoffs (NDCG@1 and NDCG@3), baseline MiniLM vs LEAP at full inference. Complements Table~\ref{tab:beir-ndcg5}; LEAP wins on ArguAna, NFCorpus, FiQA, and SciFact across cutoffs, with SCIDOCS the consistent exception.}
\label{tab:beir-ndcg1-3}
\end{table}

\section{Reproducibility}
\label{sec:appendix-reproducibility}

\paragraph{Compute Requirements.}
Training LEAP-MiniLM-L12 requires approximately 14 hours on 4$\times$NVIDIA L4 GPUs (24GB each). Single-GPU training takes proportionally longer.

\paragraph{Software.}
PyTorch 2.0+, Transformers 4.30+, sentence-transformers 2.2+.

\paragraph{Latency Measurement.}
Per-layer monitoring overhead is measured by comparing wall-clock latency of full inference (no monitoring) vs.\ full inference with convergence checking enabled but no early exits. Timing uses CUDA events with 50 warmup iterations followed by 200 measurement iterations, FP16 precision on NVIDIA L4 (24GB). All latency numbers report mean over measurement iterations.

\paragraph{Practical Adoption Checklist.}
For teams considering LEAP in production embedding pipelines, we recommend the following sequence before committing training cycles. (1) \textbf{Verify exit incompatibility}: run the diagnostic in \S\ref{sec:failure-detection} on the current distilled checkpoint to confirm that convergence-based early exit is genuinely blocked; if exit rates are already high, LEAP offers limited marginal benefit. (2) \textbf{Estimate amortization}: given measured daily embedding volume and per-request latency, estimate the number of inference hours saved per week at the target threshold $\theta$ and compare against the one-time training cost ($\sim$14 hours on 4$\times$L4). (3) \textbf{Calibrate $\theta$} on a held-out in-domain validation set — start from $\theta=0.95$ and move to $\theta=0.93$ only if quality headroom allows. (4) \textbf{Select the minimum exit layer} $l_{\min}$ based on the layer at which the baseline model first produces usable sentence representations on your domain; $l_{\min}=6$ is a reasonable default for 12-layer MiniLM-style backbones but domain-specific corpora may justify higher values. (5) \textbf{Instrument monitoring} on the exit-layer distribution and on NN@10 overlap against a full-inference reference; alert when the distribution shifts by more than one layer or when NN@10 drops by more than 20\% week-over-week. This checklist captures the operational hand-off pattern we have used internally for embedding-service onboarding.

\end{document}